\documentclass[runningheads]{llncs}
\usepackage[T1]{fontenc}
\usepackage{graphicx,verbatim}
\usepackage{booktabs}
\usepackage{hyperref}
\usepackage{amssymb}
\usepackage{amsmath}

\usepackage{caption}    
\usepackage{adjustbox}  

\usepackage{multirow}
\usepackage{bm}
\usepackage{marvosym}
\usepackage[symbol]{footmisc}

\begin{document}
\title{MMNavAgent: Multi-Magnification WSI Navigation Agent for Clinically Consistent Whole-Slide Analysis}
\titlerunning{MMNavAgent}

\author{Zhengyang Xu\textsuperscript{‡}\inst{1,3}  \and Han Li\textsuperscript{‡}\inst{1,3,4} \and Jingsong Liu\inst{1,3} \and Linrui Xie\inst{10} \and Xun Ma\inst{1} \and Xin You\inst{9} \and Shihui Zu\inst{5} \and Ayako Ito\inst{7,1} \and Xinyu Hao\inst{5,6} \and Hongming Xu\inst{5,6} \and Shaohua Kevin Zhou\inst{8} \and Nassir Navab\inst{1,3,4} \and Peter J. Schüffler \inst{1,2} \thanks{Corresponding author: \email{peter.schueffler@tum.de}} }  
\authorrunning{Anonymized Author et al.}
\institute{
Institute of Pathology, Technical University of Munich, Germany
\and
Munich Data Science Institute (MDSI), Munich, Germany
\and
Munich Center for Machine Learning (MCML), Munich, Germany
\and
Computer Aided Medical Procedures (CAMP), TU Munich, Munich, Germany
\and
Dalian University of Technology
\and
Cancer Hospital of Dalian University of Technology, Shenyang
\and
Department of Human Pathology, Juntendo University Graduate School of Medicine
\and
University of Science and Technology of China
\and
Institute of Medical Robotics, Shanghai Jiao Tong University
\and
Northwest University of China
}
\maketitle              
\footnotetext[3]{These authors contributed equally to this work.}
\begin{abstract}
Recent AI navigation approaches aim to improve Whole-Slide Image (WSI) diagnosis by modeling spatial exploration and selecting diagnostically relevant regions, yet most operate at a single fixed magnification or rely on predefined magnification traversal. In clinical practice, pathologists examine slides across multiple magnifications and selectively inspect only necessary scales, dynamically integrating global and cellular evidence in a sequential manner. This mismatch prevents existing methods from \textbf{modeling cross-magnification interactions} and \textbf{adaptive magnification selection} inherent to real diagnostic workflows.
To these, we propose a clinically consistent  Multi-Magnification WSI Navigation Agent (MMNavAgent) that explicitly models multi magnification interaction and adaptive magnification selection. Specifically, we introduce a Cross-Magnification navigation Tool (CMT) that aggregates contextual information from adjacent magnifications to enhance discriminative representations along the navigation path. We further introduce a Magnification Selection Tool (MST) that leverages memory-driven reasoning within the agent framework to enable interactive and adaptive magnification selection, mimicking the sequential decision process of pathologists.
Extensive experiments on a public dataset demonstrate improved diagnostic performance, with 1.45 \% AUC $\uparrow$ and 2.93 \% BACC $\uparrow$ over a non-agent baseline. Code will be public upon acceptance.


\end{abstract}

\section{Introduction}

\begin{figure}[!htb]
  \centering
\includegraphics[width=0.8\textwidth]{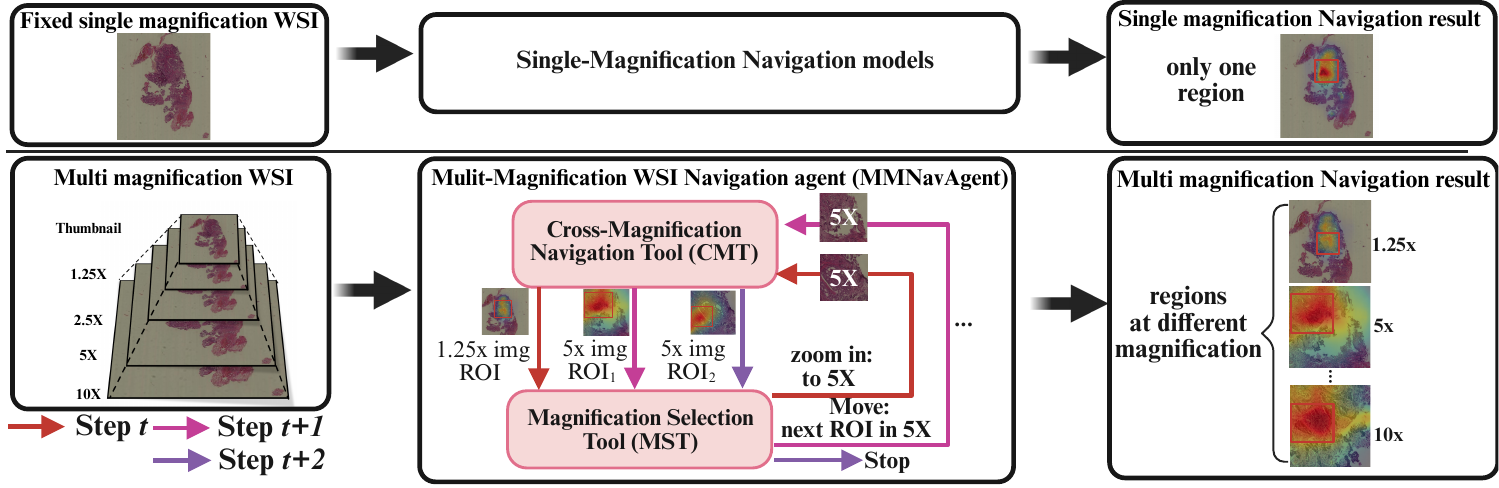}
\caption{\textbf{Top:} Prior navigation methods typically use a fixed magnification, yielding isolated single-scale cues.
\textbf{Bottom:} MMNavAgent explicitly models multi-magnification interaction and adaptive magnification selection via a closed MST-CMT loop: MST selects the next action, CMT extracts high-attention regions, and the loop repeats until Stop outputs the final multi-magnification regions.}
 \label{fig1}
\end{figure}

 AI Navigation approaches improve Whole-Slide Image (WSI) analysis by modeling spatial exploration and identifying diagnostically relevant regions\cite{raciti2023clinical,pantanowitz2020accuracy}. These regions are then incorporated into slide-level prediction frameworks, including multi-instance learning (MIL)\cite{ilse2018attention,zhao2022setmil,ding2023multi}, attention-based aggregation\cite{zhang2022whole,jiang2024multimodal}, or others\cite{nan2025deep}. Compared to conventional patch-based approaches
that treat WSIs as unordered patch collections and focus on tumor-only regions,
navigation-based methods introduce structured spatial exploration into WSI analysis\cite{nan2025deep,ghezloo2025pathfinder,sun2025cpathagent}.

However, existing approaches are typically restricted to a \textbf{fixed magnification level}.
In clinical practice, pathological WSI examination \textbf{involves dynamic navigation across multiple magnifications}\cite{thandiackal2022differentiable,agarwal2021survival}. Pathologists continuously zoom in and out, integrating tissue-level architectural context with cellular-level details to form diagnostic hypotheses\cite{ghezloo2022analysis,lee2024convolutional}. Current navigation-based methods fail to capture such cross-magnification interactions and hierarchical contextual cues, potentially limiting the effectiveness of navigation modeling and downstream diagnostic performance.

A naive solution is to independently model navigation at multiple magnifications, but this strategy still shows limitations: \textbf{(1) navigation at each scale is isolated,} whereas pathologists integrate contextual cues across magnifications; \textbf{(2)  all magnifications are exhaustively explored,} while clinicians selectively examine only the scales necessary for a given case; and \textbf{(3) magnification traversal is predefined rather than adaptive,} in contrast to the knowledge-driven and sequential decision process used in real diagnosis.

To address these, we propose a clinically consistent  \textbf{Multi-Magnification WSI Navigation Agent (MMNavAgent)} with two complementary parts (Fig.~\ref{fig1}): a \textbf{Cross-Magnification Navigation Tool (CMT)} capturing multi-scale information (limit.~1), and a \textbf{Magnification Selection Tool (MST)} for adaptive magnification selection in a clinically consistent manner (limit.~2 and~3). 

The \textbf{CMT} models cross-magnification semantic relationships. Instead of extracting features from a single magnification independently, CMT aggregates information from adjacent magnification levels, combining coarser contextual cues from lower magnifications with finer structural details from higher magnifications to enhance local feature representation. This adjacent-scale fusion reflects the hierarchical structure of pathological examination and strengthens discriminative representations along the navigation-selected path.

The \textbf{MST} governs magnification transitions within the agent. Starting with a WSI's thumbnail and a text prompt, MST determines the initial magnification level. CMT then performs navigation at the selected scale to identify diagnostically relevant regions, and the extracted representations are stored in a memory bank. Based on the stored diagnostic evidence, MST iteratively determines the next magnification action (zoom, move, or stop), and the selected scale is passed back to CMT for further exploration.

MMNavAgent coordinates MST and CMT in an iterative manner to perform structured and adaptive WSI exploration, enabling efficient magnification transitions and diagnostically relevant patch selection in a way that mirrors the sequential, knowledge-driven navigation of pathologists. Extensive quantitative and qualitative results on a public dataset \cite{nan2025deep} demonstrate its superiority and efficiency,  with gains of 1.45\% AUC and 2.93\% BACC over a non-agent baseline.

\begin{figure}[tbh]
  \centering
\includegraphics[width=0.8\textwidth]{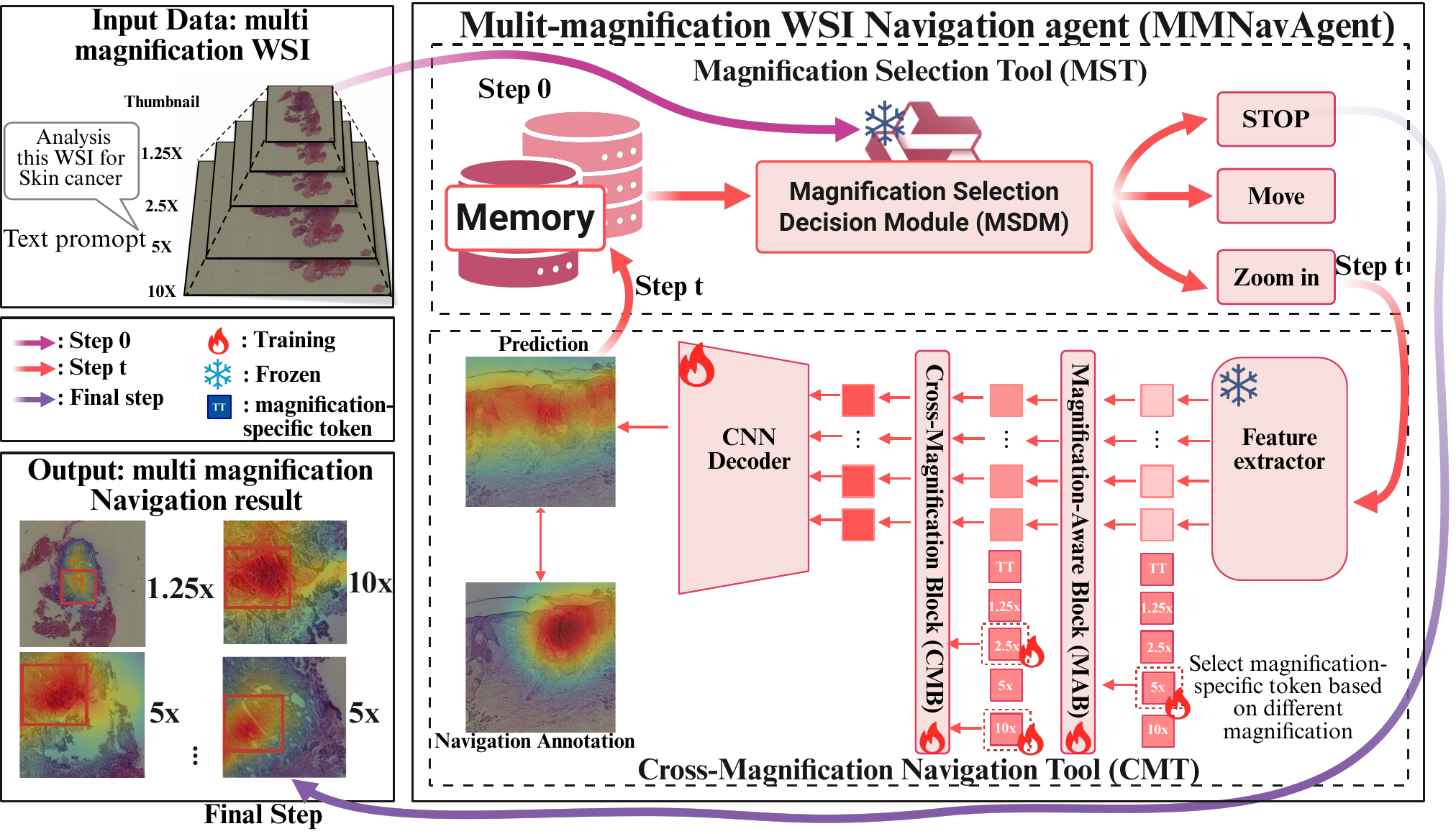}
\caption{MMNavAgent starts from thumbnail (step 0). At each step $t$, MST uses the accumulated memory to select the next action and invokes CMT to extract important regions at the chosen magnification, which are stored in memory. The loop ends (Final step) when MST selects Stop, outputting navigation results.}\label{fig2}
\end{figure}

\section{Method}
\textbf{Multi-Magnification WSI Navigation Agent (MMNavAgent)} is proposed with two parts (Fig.~\ref{fig2}):
(1) \textbf{Cross-Magnification Navigation Tool (CMT)} generates attention heatmaps across magnifications (see Sec.~2.1).
(2) \textbf{Magnification Selection Tool (MST)} iteratively and automatically invokes CMT to choose diagnostically relevant magnifications (see Sec.~2.2).
Given an thumbnail $I^{0}$ and a text prompt $T$ at initial step, MMNavAgent first determines the initial action $Op^0$ and runs iteratively.
At step $t$, MST determines the next action $Op^t$ and CMT navigates the current view to extract relevant regions $I^t$, which are then accumulated in memory $ \mathcal{M}$ for subsequent decision making.
The loop continues until sufficient evidence is collected, then outputs $\mathcal{R}$, a set of diagnostically relevant regions across magnifications:
\begin{equation}
\resizebox{0.85\linewidth}{!}{
$
\big(I^{0},\,T\big)
\xrightarrow{\text{MST}}
 Op^0
\xrightarrow{\text{CMT }}
\begin{cases}
\big(I^{1},\,T\big) & \text{if continue (zoom/move)} 
\xrightarrow{\text{CMT and MST}} \cdots \rightarrow \mathcal{R} \\
\mathcal{R}, & \text{if stop (sufficient)}
\end{cases}
$
}
\end{equation}

\subsection{Cross-Magnification Navigation Tool (CMT)}
Multi-magnification navigation faces two challenges: (1) navigation supervision is isolated across scales, making per-magnification heatmaps much sparser and harder to learn than single-scale aggregated heatmaps; and (2) eye-tracking annotations are sparse and subjective, potentially missing diagnostically important but unlabeled regions, causing ambiguous supervision.

To these, two novel designs in CMT: (1) Magnification-conditioned Fusion Network to aggregate information from adjacent magnification via cross-scale interactions to localize diagnostically relevant regions;
(2) Navigation-driven Supervision Loss to comprehensively learn diagnostic cues from sparse annotations.

\noindent\textbf{Magnification-Conditioned Fusion Network (MCFN).}
For each magnification $\mathbf{I}^m$ within the multi-scale set $\{\mathbf{I}^m\}_{m=1}^k$ of a WSI, the MCFN model leverages magnification-specific local cues and cross-magnification global context to yield a prediction heatmap $P^m$ for each level: $P^m=F_{mcfn}(I^m)$.
Specifically, the MCFN consists of four components (Fig.~\ref{fig2}): feature extractor, Magnification-Aware Block (MAB), Cross-Magnification Block (CMB), decoder.

First, each $\mathbf{I}^{m}$ is split into $N$ non-overlapping patches and encoded by a frozen pathology encoder into patch tokens
$\mathbf{X}^{m}=[\mathbf{x}^{m}_1,\mathbf{x}^{m}_2,$
$\ldots,\mathbf{x}^{m}_N]\in\mathbb{R}^{N\times D}$
, where $D$ is the token dimension. Second, the patch tokens $\mathbf{X}^{(m)}$ are feed into the MAB to obtain a \textbf{magnification-aware representation}. Inside the MAB, a learnable magnification-specific token $\mathbf{t}^{m}$ is introduced to query $\mathbf{X}^{m}$ via cross-attention, ensuring $\mathbf{X}^{m}$ are informed by the magnification context. The resulting representation is then reinjected into $\mathbf{X}^{m}$ through a gated residual connection:  
\begin{equation}
\small
\mathbf{X}^m \leftarrow \mathbf{X}^m + \gamma \cdot \text{Attn}(\mathbf{t}^m, \mathbf{X}^m, \mathbf{X}^m),
\end{equation}
where $\gamma$ is a learnable scalar gate controlling the injection strength.
Third, the updated $\mathbf{X}^{m}$ are fed into the CMB to align with \textbf{adjacent-magnification information} through spatial correspondence. Specifically, for each magnification $m$, we first derive scale-matched representations from its neighbors through spatial interpolation: a downsampled version $\hat{\mathbf{X}}^{m-1}$ from the lower level and an upsampled version $\hat{\mathbf{X}}^{m+1}$ from the higher level, ensuring that the feature dimensions are spatially aligned across adjacent levels.
These neighbor representations are subsequently refined via cross-attention with their respective magnification tokens $\mathbf{t}^{m-1}$ and $\mathbf{t}^{m+1}$, guaranteeing that the exchanged information is effectively conditioned on the target magnification level. Finally, these refined cross-magnification features are remapped to the current scale $m$ and fused into $\mathbf{X}^{m}$ via a weighted residual connection:
\begin{equation}
\resizebox{0.85\linewidth}{!}{
$ \mathbf{X}^m \leftarrow \mathbf{X}^m + u \cdot \text{Attn}(\mathbf{t}^{m-1}, \hat{\mathbf{X}}^{m-1}, \hat{\mathbf{X}}^{m-1}) + w \cdot \text{Attn}(\mathbf{t}^{m+1}, \hat{\mathbf{X}}^{m+1}, \hat{\mathbf{X}}^{m+1}), $
}
\label{eq:cross_mag_fuse}
\end{equation}
where $u$ and $w$ are learnable weights that regulate the contribution of the adjacent magnification contexts.
Finally, the updated $\mathbf{X}^m$ is feed into a simple CNN decoder $\mathcal{D}_{\text{cnn}}(\cdot)$ to produce a prediction map ${P^m}\in \mathbb{R}^{H\times W\times 1}$ for supervision with pathologist navigation heatmaps $G_{nav}^m$.

\noindent\textbf{Navigation-Driven Supervision Loss (NDSL).} 
A key limitation of navigation annotations is that the ground truth $G_{nav}^m$ is typically sparse and subjective \cite{nan2025deep}. To mitigate this, we propose a compact NDSL, which integrates a soft dice loss~\cite{milletari2016v} and a soft focal loss~\cite{lin2017focal}. This formulation encourages alignment with existing annotations while avoiding excessive penalties on unlabeled but clinically diagnostic regions. Combined with a weighted $\ell_{1}$ loss $\mathcal{L}_{1}$ as the primary supervision, the NDSL objective is mathematically formulated as:
\begin{equation}
\small
\mathcal{L}_{\mathrm{nav}}
= \lambda_{l}\,\mathcal{L}_{l}(P^m,G_{nav}^m)+\lambda_{\mathrm{dice}}\,\mathcal{L}_{\mathrm{dice}}(P^m,G_{nav}^m)
+ \lambda_{\mathrm{sig}}\,\mathcal{L}_{\mathrm{sig}}(P^m,G_{nav}^m),
\end{equation}

\subsection{Magnification Selection Tool (MST)}
In addition to multi-magnification navigation via CMT, we propose MST to enable adaptive magnification selection via two complementary components (Fig.~\ref{fig2}): 
(1) Magnification Selection Decision Module efficiently reasons over the current memory bank for the next action.
(2) Memory Bank stores MSDM decisions, accumulating case-specific pathology evidence for subsequent reasoning.
\noindent\textbf{Magnification Selection Decision Module (MSDM).}
Starting from thumbnail $\mathbf{I}^{0}$, MSDM selects an initial magnification $m$ and invokes CMT to localize $N_t$ diagnostically relevant regions $\{\mathbf{I}^{m}\}_{N_t}$, which are fed back for the next step.

Since VLMs alone are unreliable for multi-step decisions, MSDM uses a two-stage scheme:
at step $t$, the VLM describes each region selected at $t-1$, and the LLM combines these descriptions with the initial prompt $\mathbf{T}$ and accumulated memory $\mathcal{M}$ to select the next action:
\begin{equation}
\small
Op^t=
\begin{cases}
Op_{\mathrm{move}},\ \ \mathrm{LLM}(\mathrm{VLM}(\mathrm{I}),\mathrm{T},\mathcal{M}) \text{ is move}, & \\
Op_{\mathrm{zoom}},\ \ \mathrm{LLM}(\mathrm{VLM}(\mathrm{I}),\mathrm{T},\mathcal{M}) \text{ is zoom}, & \\
Op_{\mathrm{stop}},\ \ \mathrm{LLM}(\mathrm{VLM}(\mathrm{I}),\mathrm{T},\mathcal{M}) \text{ is stop}, & \\
\end{cases}\\
\label{eq:op_select}
\end{equation}
where $Op^t$ denotes the operation selected at step $t$, with ${Op_{\mathrm{move}},Op_{\mathrm{zoom}},Op_{\mathrm{stop}}}$, corresponding to \textbf{Move}, \textbf{Zoom in}, and \textbf{Stop}, respectively, as detailed below:
\textbf{(i) Move}. Stay at the current magnification $m$ and select additional high-attention, previously unselected regions output from $F_{mcfn}(\mathrm{I}^m)$.
\textbf{(ii) Zoom in}. MSDM selects a higher magnification $l>m$ and determines to select top-$k_t$ attention regions from $F_{mcfn}(I^{l})$  to zoom in.
\textbf{(iii) Stop}.When sufficient evidence is collected, output the selected multi-magnification patches.

\noindent\textbf{Memory Bank.}
The current operation $Op^t$, the selected regions $\{\mathbf{I}\}$, and their corresponding VLM-generated descriptions $\{\mathrm{VLM}(\mathbf{I})\}$ at each step are stored in memory $\mathcal{M}$, which maintains a record of all operations and provides context for subsequent decisions by MSDM: {{\footnotesize $ \mathcal{M} \;\leftarrow\; \mathcal{M}\ \cup\ \Big\{\, Op^{t},\ \{\mathbf{I}\},\ \mathrm{VLM}(\mathbf{\{I\}})\,\Big\}$}.

\section{Experiments and Results}

\subsection{Dataset and Experiment Settings}
\textbf{Dataset.} We employ the Eye-Tracking Dataset\cite{nan2025deep}, the only publicly available navigation dataset providing both pathologists’ eye-tracking  and pixel-level tumor segmentation annotations. It contains 918 skin WSI of four types: 246 nevus, 517 basal cell carcinoma (BCC), 105 melanoma, and 50 squamous cell carcinoma (SCC).
Since eye-tracking statistics show that pathologists mainly examine scales between 1.25$\times$ and 10$\times$, we generate navigation heatmaps at thumbnail, 1.25$\times$, 2.5$\times$, 5$\times$, and 10$\times$. 
We use a case-level 80\%:20\% train:test split to train CMT.

\noindent \textbf{Evaluation Protocol.}
To evaluate the effectiveness of MMNavAgent in selecting diagnostically informative regions, we assess slide-level classification performance, consistency with pathologists' navigation patterns, and tumor region overlap. For classification evaluation, we use ABMIL~\cite{ilse2018attention} as classifier and report 5-fold cross-validation results with a 60\%:20\%:20\% train:validation:test split.

\noindent \textbf{Implementation Details.}
For CMT, We adopt JWTH~\cite{liu2025linear} as the feature extractor and a U-Net~\cite{ronneberger2015u} decoder without skip connections as $\mathcal{D}_{\text{cnn}}$.
We set $\lambda_{\mathrm{l}}$ to 0.1 with a foreground weight of 2.0, while $\lambda_{\mathrm{dice}}$, and $\lambda_{\mathrm{sig}}$ are set to 1. Input size is $256 \times 256$, so MST outputs are produced at the same spatial resolution.
For MST, we use Qwen3-14B~\cite{bai2023qwen} as the LLM and Patho-R1-7B~\cite{zhang2025patho} as the VLM.

\begin{figure}[!htb]
  \centering
\includegraphics[width=0.8\textwidth]{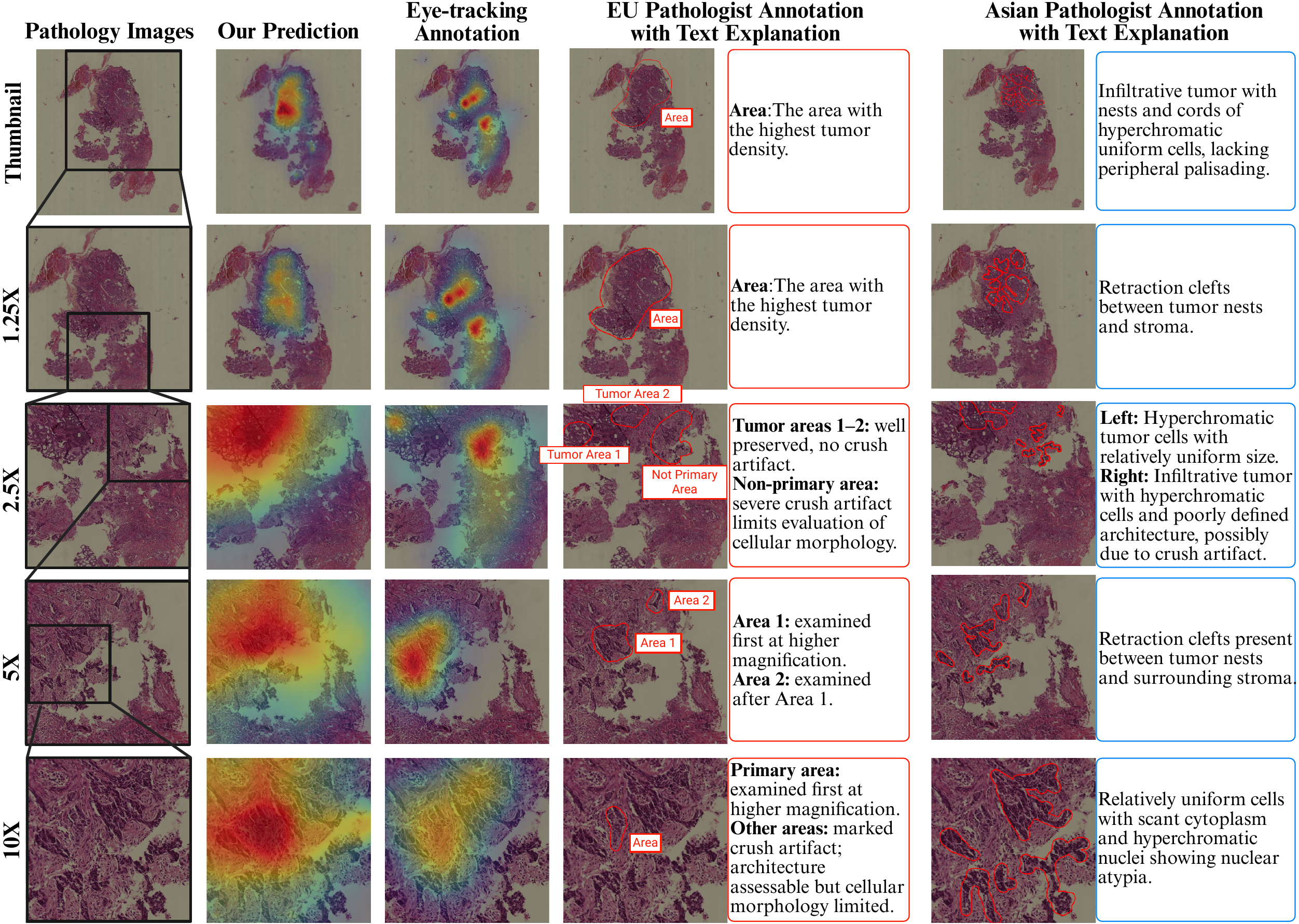}
\caption{Qualitative comparison of our predictions across five magnifications with eye-tracking annotations and two pathologists’ region annotations(EU and Asia).} \label{fig3}
\end{figure}

\begin{figure}[htbp]
\centering
\noindent
\begin{minipage}[t]{0.45\linewidth}
  \centering
  \captionof{table}{Comparison study of MMNavAgent for slide classification.}
  \label{tab1}

  \begin{adjustbox}{width=0.8\linewidth}
    \begin{tabular}{c|c|c|c}
      \hline\hline
      Method & Magnification & AUC & BACC \\
      \hline
      Baseline & fixed   & 92.88\% & 76.78\% \\
      PEAN-C\cite{nan2025deep}                 & fixed   & 93.43\% & 77.38\% \\
      PathFinder\cite{ghezloo2025pathfinder}   & fixed   & 93.13\% & 78.21\% \\
      \hline
      ours                                     & dynamic & \textbf{94.33\%} & \textbf{79.71\%} \\
      \hline\hline
    \end{tabular}
    \end{adjustbox}

    \centering
   \captionof{table}{Ablation study of MMNavAgent for slide classification.}
  \label{tab2}

  \begin{adjustbox}{width=0.75\linewidth}
    \begin{tabular}{c|c|c|c}
      \hline\hline
      Method & Magnification & AUC & BACC \\
      \hline
      ours w/o MST                             & fixed   & 94.17\% & 77.29\% \\
      ours                                     & dynamic & \textbf{94.33\%} & \textbf{79.71\%} \\
      \hline\hline
    \end{tabular}
    
  \end{adjustbox}
\end{minipage}\hfill
\begin{minipage}[t]{0.54\linewidth}
  \centering
  \captionof{figure}{Pathology consistency and tumor overlap performance (\%) for our agent}
  \label{fig4}
  \begin{adjustbox}{width=0.85\linewidth}
  \includegraphics[width=\linewidth]{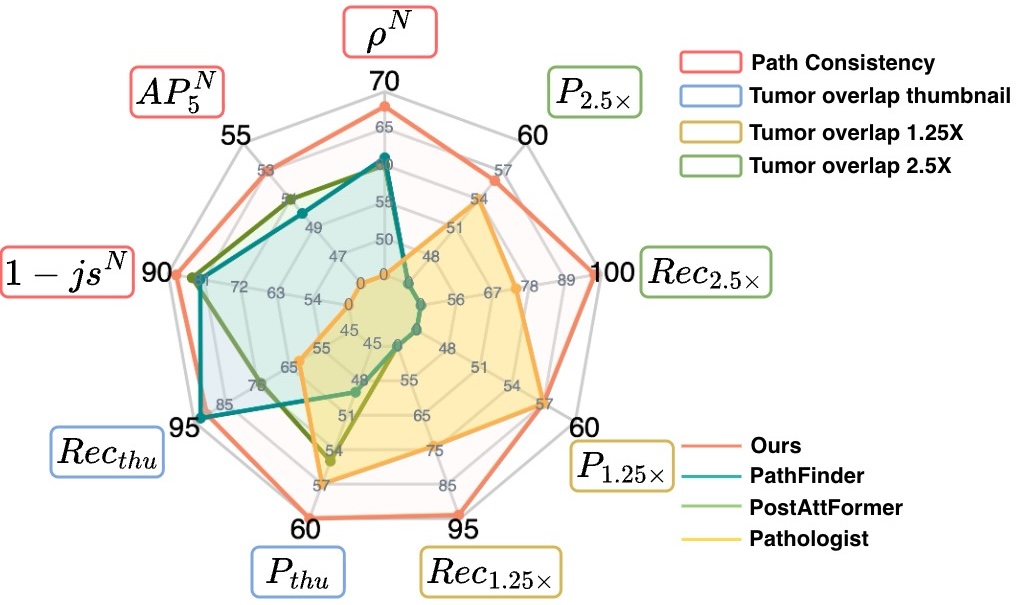}
  \end{adjustbox}
\end{minipage}
\end{figure}

\subsection{Quantitative Evaluation}
Navigation differs from segmentation: pathologists may attend to different subregions of a lesion while reaching the same diagnosis, so pixel-level overlap alone is insufficient to assess navigation consistency. 
Accordingly, we evaluate the method from three complementary perspectives: (i) slide-level classification performance, (ii) navigation consistency with pathologists, and (iii) tumor region coverage.

\noindent \textbf{Slide-Level Classification Performance.}
To assess MMNavAgent’s adaptive multi-magnification navigation, we compare it with two existing SOTA agents~\cite{nan2025deep,ghezloo2025pathfinder} on slide-level classification.
The baseline is trained on all non-background patches extracted at 10$\times$ (performs best among magnifications).
As shown in Table~\ref{tab1}, MMNavAgent achieves the \textbf{best performance} in both Area Under Curve (AUC) and Balanced Accuracy (BACC) among all methods. Its higher BACC under class imbalance suggests more reliable predictions across classes by focusing on diagnostically informative regions.

\noindent \textbf{Navigation Consistency Evaluation.}
We compare our method with two SOTA heatmap-based navigation approaches~\cite{ghezloo2025pathfinder,chakraborty2024decoding} in two ways:
(i) Agreement with eye-tracking annotations. 
Since eye-tracking annotations are subjective and spatially coarse, we quantify agreement using Spearman rank correlation $\rho^{N}$~\cite{spearman1961proof}, Average Precision with positives defined by the top-5\% ${AP}^{N}_{5}$, and Jensen–Shannon divergence $js^{N}$~\cite{lin2002divergence}. As shown in Fig.~\ref{fig4}, our method achieves the highest consistency with pathologists’ eye-tracking annotations.
(ii)  Attention Alignment with Tumor Regions.
We report ranked precision $P$ as the tumor fraction in  top-10\% attention regions (using $G_{\mathrm{tumor}}$), and compute the same metric with  eye-tracking annotations (pathologist).
Fig.~\ref{fig4} shows comparable tumor-attention proportions to pathologists across magnifications.

\noindent \textbf{Tumor Coverage Evaluation.}
To demonstrate CMT achieves comprehensive tumor coverage for more objective diagnostic evidence, we use $G_{\mathrm{tumor}}$ to compute the fraction of tumor area covered by the prediction by tumor recall $Rec$. Our method achieves the highest $Rec$ compared with pathologists and others (Fig.~\ref{fig4}).

\subsection{Qualitative Evaluation and Visualization}
We visualize predicted attention heatmaps across five magnifications alongside the eye-tracking annotations and our's additional region annotations. Two pathologists from different institutions and continents (EU and Asia) annotate these regions by marking the areas they attended to during review and adding brief textual descriptions.
As shown in Fig.~\ref{fig3}, our method aligns with eye-tracking while also highlighting additional diagnostically important areas (e.g., Area 1 at 5$\times$ annotated by the EU pathologist).
It also prioritizes clinically critical cues: at 2.5$\times$, higher attention is assigned to primary tumor-related regions rather than non-primary areas (left regions annotated by both pathologists).

\subsection{Ablation studies}

\textbf{CMT.}
We evaluated the CMT components MAB, CMB with low- (CMB-L), and high-magnification (CMB-H) with top-$k\%$  patches selected, alongside a random uniform patch-selection baseline. As shown in Table~\ref{tab3}, combining all components achieves the best AUC, highlighting the importance of each component.

\noindent \textbf{Feature extractor in CMT.}
We compare  JWTH~\cite{liu2025linear} with UNI~\cite{chen2024towards}, UNI2~\cite{chen2024towards}, CONCH1.5~\cite{lu2024visual}, and DINOv3~\cite{simeoni2025dinov3}. Using the same CMT ablation setting, JWTH performs best, as the other backbones are not optimized at the patch-token level, resulting in noisy and poorly structured heatmaps that weaken region selection. 

\noindent \textbf{NDSL loss in CMT.}
We ablate the three CMT loss terms (soft Dice loss, soft focal loss, and weighted $\ell_{1}$ loss) under the same setting as the CMT component ablation with a 40\% patch budget. Removing soft Dice or soft focal yields only small AUC drops (0.45\% and 0.47\%), whereas removing weighted $\ell_{1}$ reduces AUC by 1.23\%. This supports our design: weighted $\ell_{1}$ provides the main supervision, while the other two terms act as regularizers for unlabeled regions.

\noindent \textbf{MMNavAgent.}
To assess whether MMNavAgent can adaptively select diagnostically informative patches, we ablate MST for classification task. MST improves performance as it enables dynamic magnification selection (Table~\ref{tab2}).

\begin{table}[t]
\centering
\caption{Ablation study of CMT for slide classification (AUC of 5-fold).}
\label{tab3}
\resizebox{0.92\textwidth}{!}
{
\begin{tabular}{r|c|c|c|ccccc}
\hline
\hline

 & \multicolumn{3}{c|}{Component} & \multicolumn{5}{c}{Patch Budget}  \\
 \hline
\textbf{Sampling} &MAB &CMB-L &CMB-H  &\textbf{20\%}  &\textbf{40\%} &\textbf{60\%}  &\textbf{80\%}  &\textbf{100\%} \\
\hline
Random&&&&90.83\%$(\pm 0.74\%)$&92.24\%$(\pm 0.73\%)$&92.45\%$(\pm 0.66\%)$&92.91\%$(\pm 0.74\%)$ &92.88\%$(\pm 1.84\%)$\\
Topk&&&&91.90\%$(\pm 1.50\%)$&92.83\%$(\pm 1.52\%)$&92.47\%$(\pm 0.38\%)$&92.56\%$(\pm 0.74\%)$ &92.88\%$(\pm 1.84\%)$\\
Topk&$\checkmark$&&&91.74\%$(\pm 0.37\%)$&92.65\%$(\pm 0.94\%)$&92.78\%$(\pm 0.42\%)$&92.78\%$(\pm 0.69\%)$ &92.88\%$(\pm 1.84\%)$\\
Topk&$\checkmark$&$\checkmark$&&93.23\%$(\pm 0.82\%)$&92.70\%$(\pm 0.45\%)$&92.91\%$(\pm 0.50\%)$& 93.34\%$(\pm 0.78\%)$&92.88\%$(\pm 1.84\%)$\\
Topk&$\checkmark$&&$\checkmark$&93.02\%$(\pm 0.38\%)$&92.50\%$(\pm 0.91\%)$&92.92\%$(\pm 1.69\%)$&92.62\%$(\pm 0.69\%)$ &92.88\%$(\pm 1.84\%)$\\
Topk&$\checkmark$&$\checkmark$&$\checkmark$&\textbf{93.89\%}$(\pm 1.73\%)$&\textbf{93.29\%}$(\pm 1.12\%)$&\textbf{94.17}\%$(\pm 0.42\%)$& \textbf{93.90}\%$(\pm 0.73\%)$ &92.88\%$(\pm 1.84\%)$\\
\hline
\hline

\end{tabular}
}
\end{table}

\section{Conclusion}
In this paper, we propose a clinically consistent \textbf{Multi-Magnification WSI Navigation Agent (MMNavAgent)} with two novel components: \textbf{Cross-Magnification Navigation Tool (CMT)} models adjacent-scale interactions to capture multi-scale contextual cues, and \textbf{Magnification Selection Tool (MST)} governs adaptive magnification transitions based on accumulated diagnostic evidence in memory. Through iterative coordination between MST and CMT, MMNavAgent performs structured and adaptive WSI exploration, selecting diagnostically relevant patches without exhaustive or predefined multi-scale traversal. Extensive quantitative and qualitative results on a public dataset demonstrate improved slide-level classification with fewer, higher-quality patches, while producing navigation behaviors that better align with clinical practice.\\

\noindent\textbf{Acknowledgments.}This work was supported by th SATURN3 project (01KD2206C) and the IMI BIGPICTURE project (IMI945358).

\bibliographystyle{splncs04}
\bibliography{main.bib}

\end{document}